\pdfoutput=1
\documentclass[11pt]{article}

\usepackage[]{ACL2023}

\usepackage{times}
\usepackage{latexsym}
\usepackage{tabularx}
\usepackage{multirow}
\usepackage[T1]{fontenc}
\usepackage[utf8]{inputenc}

\usepackage{microtype}

\usepackage{inconsolata}

\usepackage{multirow}
\usepackage{multicol}
\usepackage{graphicx}
\title{SSS at SemEval-2023 Task 10: Explainable Detection of Online Sexism using Majority Voted Fine-Tuned Transformers
}

 \author{Sriya Rallabandi \hspace{1cm} Sanchit Singhal  \hspace{1cm}  Pratinav Seth \\
         Manipal Institute of Technology\\ Manipal Academy of Higher Education, Manipal, India\\
 \texttt{\{sriyarallabandi,sanchitsinghal,seth.pratinav\}@gmail.com}}
\begin{document}
\maketitle
\begin{abstract}
This paper describes our submission to Task 10 at SemEval 2023-Explainable Detection of Online Sexism (EDOS) \citet{Kirk2023SemEval2023T1}, divided into three subtasks. The recent rise in social media platforms has seen an increase in disproportionate levels of sexism experienced by women on social media platforms. This has made detecting and explaining online sexist content more important than ever to make social media safer and more accessible for women. Our approach consists of experimenting and finetuning BERT-based models and using a Majority Voting ensemble model that outperforms individual baseline model scores. Our system achieves a macro F1 score of 0.8392 for Task A, 0.6092 for Task B and 0.4319 for Task C and the code is available on GitHub.\footnote{\url{https://github.com/sriya26/SemEval}}
\end{abstract}

\section{Introduction}
Social media has become a free and powerful tool for communication, allowing people to share their ideas and connect with others worldwide. For women, social media has provided a platform for bringing visibility, empowering, and educating people on women’s issues. However, with an increase in the number of users, there is a significant increase in sexism on these platforms has led to online harassment and disinformation, making social media less accessible and unwelcoming, continuing to uphold unequal power dynamics and unfair treatment within society. Social media’s anonymity and ease of access have made it easier for individuals to spread sexist content online. This has led to a growing interest and needs to detect and identify online content as sexist and explain why that content is sexist. Natural Language Processing (NLP) techniques have shown promise in enabling the automated detection of sexist language and providing insights into its nature of it.\\
This paper describes our contributions to SemEval 2023 Task 10:Explainable Detection of Online Sexism \citet{Kirk2023SemEval2023T1}, which proposed a new dataset of sexist content from Gab and Reddit to detect and explain online sexist content through three hierarchical classification subtasks.\\
Task 10 of SemEval 2023 includes three hierarchical subtasks: Task A, which is a binary classification task to classify whether a post is sexist or not sexist, Task B, a multi-class classification task for sexist posts which further classifies them into four categories and Task C, another multi-class classification task to classify the sexist posts into one of eleven fine grained vectors. We conducted detailed experiments on BERT-based models, such as RoBERTa \citet{DBLP:journals/corr/abs-1907-11692} and DeBERTaV3 V3 \citet{DBLP:journals/corr/abs-2111-09543}, using different training methods and loss functions to reduce class imbalance along with hyperparameter tuning. We then ensembled these individual models using a majority voting ensemble and show how these methods affect the model's performance and how an ensemble of these models performs better than the individual baselines.

\section{Background}
\subsection{Problem and Data Description}
SemEval 2023 Task 10: Explainable Detection of Online Sexism aims to detect and flag any abuse or negative sentiment towards women based on their gender or based on their gender combined with one or more other identity attributes.
The dataset consists of 20,000 labelled entries. Of these, 10,000 are sampled from Gab and 10,000 from Reddit. Initially, three skilled annotators assign labels to all entries, and any discrepancies are resolved by one of the two experts. In Task A, if all annotators unanimously agree on a label, it is considered the final label. However, if there is any disagreement, an expert will review the entry and determine the final label. For Task B and C, if at least two annotators agree on a label, that is considered the final label. However, an expert will determine the label if there is a three-way disagreement.
The challenge is divided into three  hierarchical subtasks:\\
\textbf{Subtask A: Binary Sexism Classification}- The task is a two-class (binary) classification that detects whether a post is sexist. The training set consists of 14000 (70\% split) entries, of which 3398 are sexist, as shown in Table \ref{tab:A}.
\begin{table}[h]
  \centering
  \begin{tabular}{|p{5cm}|p{2cm}|}
    \hline
    Post & Label \\
    \hline
    yeah but in order to keep the benefit i have to be good tommorow because i told her we could try & not sexist\\
    \hline
    As Roosh said having discussions with woman is a waste of time anyway & sexist \\
    \hline
  \end{tabular}
  \caption{Examples for Task A}
  \label{tab:A}
\end{table}

\textbf{Subtask B: Category of Sexism}-  The task is a multi-class classification where we have to predict the sexist class into four categories (1) threats, (2) 
\begin{table}[h]
  \centering
  \begin{tabular}{|p{5cm}|p{2cm}|}
    \hline
    Post & Label \\
    \hline
    Shove PORK down her THROAT! & 1.threats,plans to harm and incitement\\
    \hline
    Classic toxic white females & 2.derogation \\
    \hline
    Rule one: No woman can be incel, only volcel. & 3. animosity \\
    \hline
    English: robot know best person for job. Robot no choose woman. Robot choose man.& 4. prejudiced discussions \\
    \hline
  \end{tabular}
  \caption{Examples for Task B categories}
  \label{tab:B}
\end{table} 
derogation, (3) animosity, (4) prejudiced discussions, shown above in Table \ref{tab:B}.\\
\begin{figure}[h]
    \centering
    \includegraphics[width=1\linewidth]{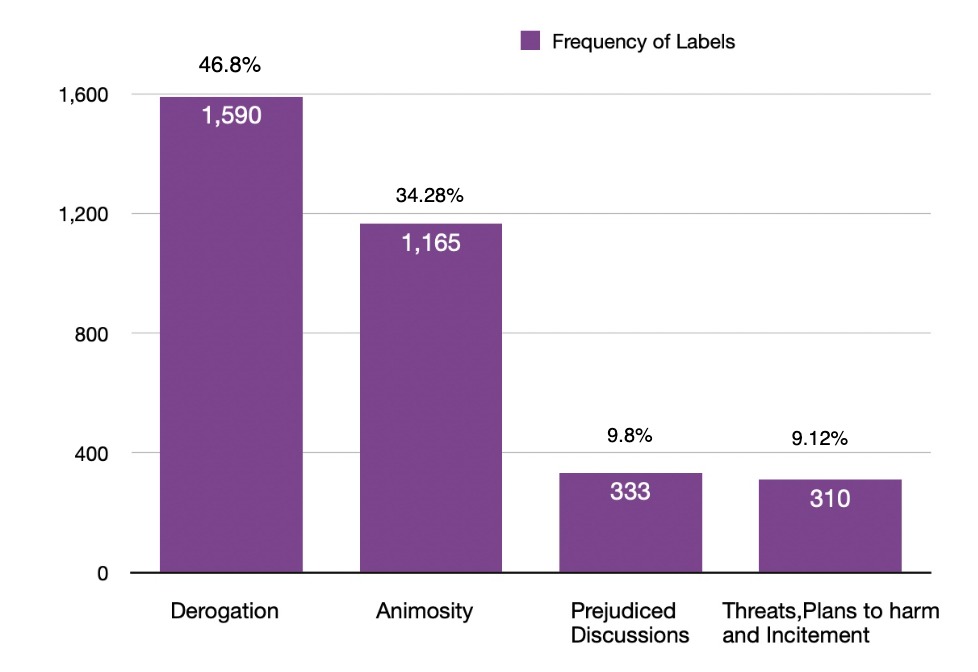}
    \caption{Frequencies of Task B labels in training set}
    \label{fig:B}
\end{figure}\\
\textbf{Subtask C: Fine-grained Vector of Sexism}- The task is a multi-class classification where the sexist posts are further subdivided into 11 fine-grained Vector of Sexism: (1.1) Threats of harm and Incitement and (1.2) encouragement of harm are included in the Category of threats ; (2.1) Descriptive attacks, (2.2) Aggressive and emotive attacks and (2.3) Dehumanising attacks and overt sexual objectification in Derogation ; (3.1) Causal use of gendered slurs, profanities and insults, (3.2) Immutable gender differences and gender stereotypes,(3.3) Backhanded gendered compliments and (3.4) Condescending explanations or unwelcome advice in Animosity; (4.1) Supporting mistreatment of individual women and (4.2) Supporting systemic discrimination against women as a group in Prejudiced discussions, as shown in Table \ref{tab:C}.\\
\begin{figure}[h]
    \centering
    \includegraphics[width=1\linewidth]{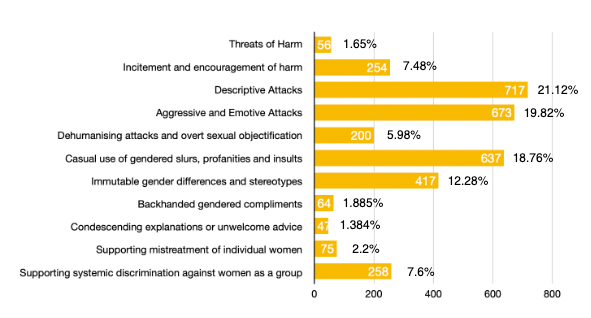}
    \caption{Frequencies of Task C labels in training set}
    \label{fig:B}
\end{figure}
\begin{table}[h]
  \centering
  \begin{tabular}{|p{5cm}|p{2cm}|}
    \hline
    Post & Label \\
    \hline
    Shove PORK down her THROAT! & 1.2 incitement and encouragement of harm \\
    \hline
     Classic toxic white females& 2.1 descriptive attacks \\
    \hline
    Rule one: No woman can be incel, only volcel.& 3.2 immutable gender differences and gender stereotypes \\
    \hline
    English: robot know best person for job. Robot no choose woman. Robot choose man.& 4.2 supporting systemic discrimination against women as a group\\
    \hline
  \end{tabular}
  \caption{Examples for Task C vectors}
  \label{tab:C}
\end{table}
\subsection{Related Work}
\subsubsection{Onine Sexism}
\textbf{Online Sexism }is considered a type of hate speech by many and can be found on several big platforms like Twitter and Reddit. This increases the importance of improving detection and classification accuracy for the same. Major work has been done in Harmful Speech Detection in previous SemEval tasks like Hate news detection \citet{basile-etal-2019-semeval}; Propaganda Detection \citet{da-san-martino-etal-2020-semeval}; and Rumor \citet{gorrell-etal-2019-semeval} Detection.
\citet{stanczak2021survey} conducted a survey\\
\citet{fersini-etal-2022-semeval} aimed to detect misogynous memes on the web by taking advantage of available texts and images, and \citet{mina2021automatic} and \citet{rodriguez2020automatic} aims to automatically identify sexism in social media content by applying machine learning methods.
\subsubsection{Methods for detecting Online Sexism}
\citet{de2022detection} explored the use of different transformers architectures for two tasks in English and Spanish: sexism detection and sexism classification.
\citet{parikh2019multi} introduced the first work on multi-label classification for sexism detection and also provided the largest dataset on sexism categorization. They built a BERT-based neural architecture with distributional and word-level embeddings to perform the classification task.\citet{}
\citet{abburi2021knowledge} proposed a neural model for this sexism detection and classification that can combine representations obtained using RoBERTa model and linguistic features such as Empath, Hurtlex, and PerspectiveAPI by involving recurrent components.
\citet{agrawal-mamidi-2022-lastresort} and \citet{rao-2022-asrtrans} performed binary and multi-class classification using transformer-based models and showed how ensembling them outperforms the baseline model performances.

\section{System Overview}
After conducting extensive experiments on pretrained transformer-based models such as BERT \citet{DBLP:journals/corr/abs-1810-04805}, RoBERTa \citet{DBLP:journals/corr/abs-1907-11692}, DeBERTa \citet{DBLP:journals/corr/abs-2006-03654},DeBERTaV3 \citet{DBLP:journals/corr/abs-2111-09543}, AlBERTa \citet{DBLP:journals/corr/abs-1909-11942} and XL-Net \citet{DBLP:journals/corr/abs-1906-08237} and comparing their baseline performances based on accuracy and Macro F1 score, we finally used DeBERTaV3 for Task A and both RoBERTa and DeBERTaV3 for Task B and C.
\subsection{Ensemble}
To increase the overall accuracy of the predictions and robustness of the predictive model, models were first individually tuned on the entire dataset. Then, their predictions were combined using ensembling methods. Final results were obtained using the Majority Voting Ensembling method, a hard voting method in which the class predicted by most models is considered the final output and weighted average ensemble, in which the weights are obtained by grid search or other optimization techniques on the validation dataset. 

\subsection{Task A: Binary Classification}

This consisted of classifying texts into binary classes: sexist and not sexist.
The data was trained on the pre-trained BERT-based models, of which DeBERTaV3 gave the best baseline score. We experimented with the DeBERTaV3  model with tuning parameters such as different learning rates and number of epochs to test their effect on the model's accuracy. We also used Focal Loss \citet{DBLP:journals/corr/abs-1708-02002} and Class Weights to deal with class imbalance in the dataset. We finally used the Majority Voting ensembling method on the top 3 model combination results to obtain the final labels.

\subsubsection{Class Imbalance}
We noticed a major class imbalance in the training dataset of Task A, with 10602, not sexist examples and only 3398 sexist examples. Hence to deal with this, we experimented with class imbalance handling measures such as Focal Loss and Class Weights assigned to each class to reduce class imbalance. \\
\textbf{Focal Loss:} Focal Loss is primarily used in classification tasks to deal with class imbalance. 
\begin{equation}
    L_{FL}= \alpha(1 - p)^\gamma \log(p)
\end{equation}
where $\gamma$ and $\alpha$ are the hyperparameters. 

It is an improvement over the Cross-Entropy loss function.\citet{DBLP:journals/corr/abs-1805-07836} and deals with class imbalance by assigning more weights to hard or easily misclassified examples and down weight easy examples, thereby emphasizing correcting misclassified ones.$\gamma$>1 will rescale the modulating factor such that the easy examples are down-weighted more than the hard ones, reducing their impact on the loss function.\\

\subsection{Task B and C: Multi-Label Classification}
These tasks were a multi-label classification problem. We had to predict the category label in task B and the vector label in task C of the text already classified as sexist in task A. We experimented with parameters like learning rates and the number of epochs. We used Focal Loss and Class Weights for any class imbalance. Apart from these methods, we also trained the baseline DeBERTaV3  and RoBERTa models using a 5-fold cross-validation method and compared the effect of the training method on accuracy. The best-performing models were ensembled using the Majority Voting ensemble method to get the final labels.

\section{Experimental Setup}
The training data consists of 14,000 entries out of 20,000 total entries, and these are used for Task A. 3,398 of these 14,000 are labelled as sexist, which are further used in Tasks B and C. 
We strictly used the data provided by the organisers to train and test our models in the development and testing phases. We used 80\% of the training dataset to train the models, while 20\% remaining was used for cross-validation. 
Upon using both methods, we found that the majority voting ensemble method gives better results than the weighted average ensemble method. Using the majority voting ensemble reduced individual model errors, and the impact of outliers was also minimized, leading to a more accurate and robust system.\\
For Task A, we used our best three performing models for the ensemble, including the baseline DeBERTaV3  model, with focal loss and different hyperparameter values. The exact hypermeters used for these three models are mentioned in Table \ref{tab:hyp}.
\begin{table}[h]
  \centering
  \begin{tabular}{|c|c|c|c|}
    \hline
    \textbf{Parameters} & \textbf{Model 1} & \textbf{Model 2} & \textbf{Model 3} \\
    \hline
    Learning Rate & 2e-5 & 2e-5 & 4e-5 \\
    Epochs & 6 & 6 &  4 \\
    Batch Size & 6 & 6 & 6 \\
    Loss Function & None & Focal & Focal  \\
    Optimizer & None & Adam & Adam \\
    \hline
  \end{tabular}
  \caption{Hyperparameters for Task A DeBERTaV3 models}
  \label{tab:hyp}
\end{table}\\
Model 1 is the baseline DeBERTaV3 model with a learning rate of 2e-5 and batch size and epochs of 6. Model 2 is the baseline as Model 1 but with Focal loss and Adam optimizer while Model 3 has a learning rate of 4e-5 and runs on 4 epochs instead.\\
 \(\alpha\) and \(\gamma\) are the hyper-parameters of Focal Loss, where \(\alpha\) is a weighting factor that gives more importance to the minority class, and \(\gamma\) is a tunable parameter that modulates the degree of down-weighting. The values we take after experimenting are 1.0 for \(\alpha\) and 2.0 for \(\gamma\), the default value used in the original paper.
We also made use of Class Weights along with Focal Loss, which are computed for each class using the formula,
\begin{equation}
    \frac{ n_{samples} }{ n_{classes} \times {np.bincount}(y)}
\end{equation}
However upon experimenting, we observed that the results obtained were lesser compared to baseline models, hence we did not include it in the final ensemble.\\
In Task B and C, after much experimenting, we noticed that an ensemble of RoBERTa and DeBERTaV3  models with 5-fold cross-validation gave a better result than the baseline models. The learning rate for both was taken as 2e-5, with 4 epochs in each fold.\\
Macro F1 score was the primary evaluation metric used. We used the pretrained models available in huggingface library and implemented the models using ktrain library \citet{maiya2020ktrain}, a lightweight wrapper for deep learning libraries.
\section{Results}
Results for Task A, B and C are mentioned in Tables \ref{tab:results} and \ref{tab:test} as F1 scores. We have presented our results from the Development phase in Table \ref{tab:results} and the final results of the ensemble from the Testing phase in Table \ref{tab:test}. We ranked 34th out of 84 for Task A, 36th out of 69 for Task B and 30th out of 63 for Task C. 
\begin{table}[h]
  \centering
  \begin{tabular}{|l|l|l|}
    \hline
    \textbf{Task} & \textbf{F1 Score}  \\
    \hline
    Task A & 0.8392 \\ 
    \hline
    Task B & 0.6092 \\ 
    \hline
    Task C & 0.4319 \\
    \hline  
  \end{tabular}
  \caption{Results on Test Phase dataset with Ensemble}
  \label{tab:test}
\end{table}

\begin{table}[htbp]
    \centering
    \begin{tabular}{|p{0.9cm}|p{4cm}|p{1.4cm}|}
       \hline
       \textbf{Task} & \textbf{Model} & \textbf{F1 Score}  \\
       \hline
       \multirow{3}{2.5cm}{Task A} & DeBERTaV3  & 0.8348 \\ 
        & DeBERTaV3 {with focal loss} & 0.8346 \\ 
        & DeBERTaV3 {with lr=4e-5} & 0.8345 \\ 
       & \textbf{Ensemble}  & \textbf{0.8411} \\
       \hline
       \multirow{3}{2.5cm}{Task B} & DeBERTaV3  & 0.6381 \\ 
       & RoBERTa & 0.6376 \\
       & DeBERTaV3 {with 5-fold CV} & 0.6522 \\ 
       & RoBERTa {with 5-fold CV} & 0.6490 \\  
       & \textbf{Ensemble}  & \textbf{0.6592} \\
       \hline
       \multirow{3}{2.5cm}{Task C} & DeBERTaV3  & 0.4285 \\
       & RoBERTa & 0.4263 \\
       & DeBERTaV3 {with 5-fold CV} & 0.4336 \\ 
       & RoBERTa {with 5-fold CV} & 0.4327 \\
       &\textbf{Ensemble}  & \textbf{0.4469} \\
       \hline   
    \end{tabular}
  \caption{Results for Development phase}
  \label{tab:results}
\end{table}

While using focal loss doesn't improve the F1 score compared to the baseline DeBERTaV3  model in Task A, ensembling both the models along with DeBERTaV3  with a learning rate of 4e-5 improves the overall performance of the model.

For Task B and C, Focal Loss models performed poorly compared to the baselines despite having some imbalanced classes. We also noticed that using a different training method, like 5-fold Cross Validation improved the performance over the baseline models for multi-class classification and ensembling them further improved the performance.

\section{Conclusion and Future Works}
\label{sec:bibtex}

In this work, we benchmarked various pre-trained Transformer based models such as BERT, RoBERTa and DeBERTaV3  and their majority voting Ensemble Models. This detection and classification of online sexism were done under Task 10 at SemEval 2023-Explainable Detection of Online Sexism (EDOS). We also used Focal loss to deal with class imbalance.DeBERTaV3  gave the best F1 score of 0.8348 on Task A, 0.6381 on Task B, and 0.6381 on Task C. RoBERTa gave results close to DeBERTaV3  on both Task B and C.

In future works, we would like to \textbf{(1)} compare the results of the models on different datasets and languages; \textbf{(2)} use unsupervised learning to improve the results of our models;\textbf{(3)} propose a novel custom architecture which provides more robust predictions.

% Entries for the entire Anthology, followed by custom entries
\bibliography{anthology,custom}
\bibliographystyle{acl_natbib}

\end{document}